%% file: main.tex
\documentclass[10pt,twocolumn,letterpaper]{article}
\usepackage{cvpr}      
\usepackage{multirow}    
\usepackage{arydshln}
\usepackage{lineno}
\usepackage{xcolor}       
\usepackage{colortbl}     
\usepackage{multirow}     
\usepackage{arydshln}     
\usepackage{amsmath}      
\usepackage{caption}
\usepackage{booktabs}    
\usepackage{makecell}    
\usepackage{multirow}    

\definecolor{cvprblue}{rgb}{0.21,0.49,0.74}
\usepackage[pagebackref,breaklinks,colorlinks,allcolors=cvprblue]{hyperref}
\def\paperID{} 
\def\confName{CVPR}
\def\confYear{2026}

\title{MRI Contrast Enhancement Kinetics World Model}
\author{
Jindi Kong$^{1}$ \quad
Yuting He$^{1}$ \quad
Cong Xia$^{2}$ \quad
Rongjun Ge$^{3}$ \quad
Shuo Li$^{1}$\thanks{Corresponding author: shuo.li11@case.edu}\\
$^{1}$Case Western Reserve University, Cleveland, OH, USA,\\ $^{2}$Jiangsu Cancer Hospital, Nanjing, Jiangsu, China, $^{3}$Southeast University, Nanjing, Jiangsu, China\\[0.6em]
}

\begin{document}
\maketitle
\input{abstract}    
\input{intro}
\input{Related_Work}
\input{Method}

\input{Experiments}

\input{Conclusion}
{
    \small
    \bibliographystyle{ieeenat_fullname}
    \bibliography{main}
}
\input{X_suppl}

\end{document}

%% file: abstract.tex
\begin{abstract}
Clinical MRI contrast acquisition suffers from inefficient information yield, which presents as a mismatch between the risky and costly acquisition protocol and the fixed and sparse acquisition sequence. Applying world models to simulate the contrast enhancement kinetics in the human body enables continuous contrast-free dynamics. However, the low temporal resolution in MRI acquisition restricts the training of world models, leading to a sparsely sampled dataset. Directly training a generative model to capture the kinetics leads to two limitations: (a) Due to the absence of data on missing time, the model tends to overfit to irrelevant features, leading to content distortion. (b) Due to the lack of continuous temporal supervision, the model fails to learn the continuous kinetics law over time,  causing temporal discontinuities. For the first time, we propose \textbf{MRI} \textbf{C}ontrast \textbf{E}nhancement \textbf{K}inetics \textbf{World} model (MRI CEKWorld) with \textbf{S}patio\textbf{T}emporal \textbf{C}onsistency \textbf{L}earning (STCL). For (a), guided by the spatial law that patient-level structures remain consistent during enhancement, we propose Latent Alignment Learning (LAL) that constructs a patient-specific template to constrain contents to align with this template. For (b), guided by the temporal law that the kinetics follow a consistent smooth trend, we propose Latent Difference Learning (LDL) which extends the unobserved intervals by interpolation and constrains smooth variations in the latent space among interpolated sequences. Extensive experiments on two datasets show our MRI CEKWorld achieves better realistic contents and kinetics. Codes will be available at https://github.com/DD0922/MRI-Contrast-Enhancement-Kinetics-World-Model.
\end{abstract}

%% file: intro.tex
\section{Introduction}
\label{sec:intro}
\begin{figure}[tbp]
    \centering
    \includegraphics[width=\linewidth]{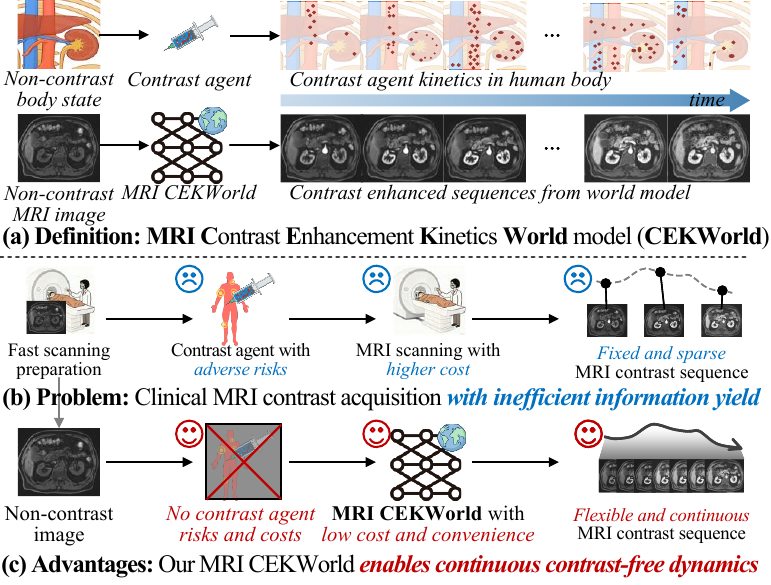} 
    \caption{(a) \textbf{Task}: MRI CEKWorld generates contrast-enhanced sequences that conform to kinetics in the human body after contrast agent injection. (b) \textbf{Problem}: Clinical contrast MRI acquisition presents inefficient information yield with adverse risks and higher cost, but a fixed, sparse sequence.  (c) \textbf{Advantages}: Our MRI CEKWorld enables continuous contrast-free dynamics with no contrast agent risks, low cost, and convenience.}
    \label{fig:def}
\end{figure}
World models \cite{ding2024understanding, lecun2022path, ha2018world}, which learn to simulate the dynamics of physical systems via deep neural representations \cite{ cheng2024pixelwise,Murugesan312,muller2024diffusion,osuala2025simulating}, offer a compelling direction for modeling MRI contrast enhancement kinetics. As illustrated in Fig.~\ref{fig:def}, such models can infer the pharmacokinetic evolution of contrast agents directly from an initial non-contrast MRI, thereby enabling the estimation of contrast-agent distribution at arbitrary time points and synthesizing the corresponding contrast-enhanced MRI images. Once realized, this capability yields two key advantages: \textbf{1)} \textit{Contrast-free MR imaging paradigm.} By obviating the need for exogenous contrast administration, it mitigates injection-related risks \cite{muller2023using} and reduces the economic \cite{shankar2018financial} and procedural overhead incurred by additional contrast-enhanced acquisitions. \textbf{2)} \textit{High temporal resolution modeling.} By producing continuous and temporally dense enhancement trajectories which are unconstrained by the sparse sampling of clinical protocols \cite{strijkers2007mri,ingrisch2013tracer}, the model offers substantially higher information throughput and a more faithful reconstruction of underlying contrast-agent kinetics. These considerations naturally motivate a scientific question: ``\textit{Can we construct an MRI \textbf{C}ontrast \textbf{E}nhancement \textbf{K}inetics \textbf{World} model (CEKWorld) that, relying solely on non-contrast MRI, faithfully reconstructs in vivo contrast-agent dynamics and produces high-fidelity temporal enhancement images?}"

However, it is challenging to train the MRI CEKWorld owing to the low temporal resolution in MRI acquisition \cite{okada2022dreamingv2,gao2024vista,hafner2019dream}. Different from the general domain where millisecond-level continuous videos can be obtained via continuous sampling for model training \cite{hollingsworth2015reducing,andria2009acquisition,plewes2012physics,kang2024far}, the sequence acquisition in the MRI CEKWorld is constrained by reconstruction duration \cite{andria2009acquisition,hollingsworth2015reducing, plewes2012physics} and patient respiratory cooperation \cite{smith2010mri}. Consequently, the acquired sequences are extremely sparse, with only second-level intervals \cite{yamada2022time}, such a sparsely sampled dataset directly hinders the model's learning of contrast agent kinetic laws.

\begin{figure}[tbp]
    \centering 
    \includegraphics[width=0.45\textwidth]{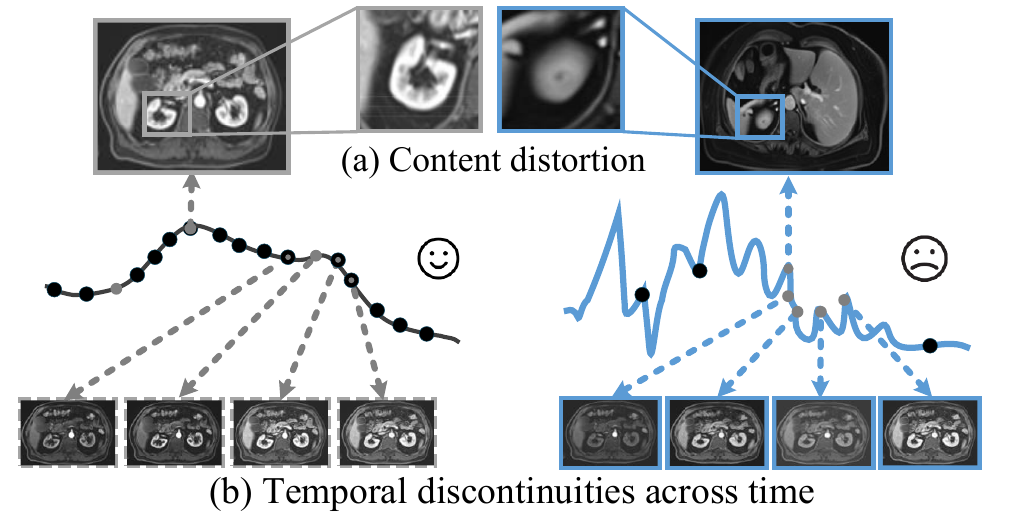} 
    \caption{\textbf{Limitation}: MRI Acquisition-induced low temporal resolution in MRI CEKWorld leads to (a) Content distortion and (b) Temporal discontinuities across time.}
    \label{fig:challenge} 
\end{figure}

A practical compromise is to directly train a generative model on a sparsely sampled dataset in an attempt to capture the underlying contrast-agent kinetics. However, this strategy suffers from two fundamental limitations: (a) Content distortion in the spatial dimension. Owing to the absence of ground-truth frames at the missing time points, the model receives no supervision on the true anatomical state. Once overfitting occurs, it produces the distortions illustrated in Fig.~\ref{fig:challenge}(a), including structural deformation and organ misalignment. Although prior-based regularization \cite{huang2021medical, akbari2024beas, phan2023structure, ma2021structure, zotti2018convolutional} can encourage more realistic content at unsampled times, such priors still fail to preserve patient-specific anatomical details. (b) Discontinuity in the temporal dimension. Without continuously sampled data, the model is unable to learn the true kinetic law of the contrast agent, leading to mismatches with time conditions and temporal jumps between adjacent frames, as shown in Fig.~\ref{fig:challenge}(b). While post-hoc smoothing \cite{jeong2023deep,Guo_2024_CVPR,10.1145/3664647.3680846,mallet1989discrete} reduces visible discontinuities, pixel-space smoothing inevitably blurs fine details and deviates from the actual kinetics.

The contrast kinetics of MRI in the same patient follow an inherent spatiotemporal consistent law. \textit{Spatially}, the anatomical structures such as organ contours, tissue boundaries, and their relative positions in the same patient remain consistent across time, unaffected by the dynamic changes of contrast agent kinetics. Guided by this spatial law, patient-level spatial consistency is enforced to constrain the model’s learning, directing it to focus on relevant spatial features, effectively preserving content reality. \textit{Temporally}, the enhanced sequences follow a consistent smooth evolutionary trend without abrupt jumps. Leveraging this temporal law directs the model to capture the inherent sequential dynamics of contrast agent metabolism in the latent space, ensuring the temporal smoothness of generated sequences and suppressing unnatural jumps.

\textit{For the first time}, we propose the \textbf{S}patio\textbf{T}emporal \textbf{C}onsistency Learning (STCL), which utilizes the inherent consistency spatiotemporal law of contrast agent kinetics to enable the MRI CEKWorld, achieving realistic predictions and smooth simulations under the training from the sparsely sampled dataset. It has two innovations:

Latent Alignment Learning (LAL) for realistic content automatically constructs an explicit patient-specific template by leveraging region-specific responses to encode spatial consistent relationships and constraining the generated content at each time point to align with this template spatially. First, a patient-specific template is calculated by computing covariance matrices between features at each time point in the latent space, which represents the time-invariant spatial anatomical structure during enhancement process. It then normalizes and aggregates these features to form an explicit patient-level template. Subsequently, the equidistance constraint aligns the statistical features at each time point with this template, ensuring all time points adhere to the unified statistical rules of the template and thereby maintaining content consistency.

Latent Difference Learning (LDL) for temporal continuity interpolates in the unobserved intervals in the latent space and constrains smooth variations between consecutive points for semantic continuity. First, it uniformly inserts intermediate virtual time points between the original sparse acquisition sequence to construct a dense sequence, filling temporal gaps. Second, it calculates the variations of adjacent time points by computing the discrete second-order central differences of time points in the dense sequence, and constrains the variation to zero, which constrains the temporal changes to suppress abrupt jumps, and ensures the smoothness of temporal evolution.

Our contributions are summarized as follows: 1) For the first time, we propose MRI CEKWorld, which simulates the contrast agent kinetics in the human body and facilitates continuous contrast-free dynamics. 2) We propose STCL, which enforces content reality and temporal continuity under acquisition-induced low temporal resolution through spatiotemporal consistent physiological law. 3) Our LAL constructs an explicit patient-level template for each generation alignment, maintaining content consistency and reality. 4) Our LDL extends unobserved intervals and constrains variations between consecutive points, thereby ensuring the smoothness of temporal evolution.

%% file: Related_Work.tex
\section{Related Work}
\label{sec:formatting}

\textbf{Virtual MRI Contrast Enhancement} is an alternative to the use of contrast agents \cite{kleesiek2019can, cheng2024pixelwise, lindner2021virtual, kim2022tumor}, designed to emulate the visibility of specific tissues and bodily fluids. Compared to traditional contrast agents \cite{hayes2002assessing}, it has three significant advantages: i.) Safe: it can address safety concerns regarding possible contrast agent deposition \cite{muller2023using}, ii.) Comfortable: it can mitigate patient discomfort while scanning \cite{nitta2024relationship}, iii.) Economical: it can help cut down on human resources, hardware costs, and overall expense \cite{shankar2018financial}.

Recent virtual MRI contrast enhancement methods are divided into static generation and dynamic sequence generation.
Static generation methods \cite{cheng2024pixelwise, chen2022synthesizing, murugesan2024synthesizing, huang2025deep}  focus on synthesizing the single-phase contrast-enhanced images at a single time point,  T1-weighted contrast-enhanced images from multiple non-contrast sequences such as T1-weighted, T2-weighted and the Apparent Diffusion Coefficient map. Such methods prioritize accurately simulating the final enhancement patterns of tumors or lesions to improve the accuracy of classification-based diagnosis.
Dynamic sequence generation methods \cite{osuala2024pre, ramanarayanan2025dcetriformer, sadhana2024dce} aims to synthesize the time sequence, including multi-phase contrast-enhanced images at several time points.
However, due to the physical acquisition limit, virtual MRI contrast agent remains confined to the scope of image-to-image mapping\cite{ramanarayanan2024dce, li2024image, osuala2025simulating}, failing to simulate contrast agent kinetics.

\textbf{World Models} are able to understand the world and predict the future \cite{ding2024understanding, lecun2022path, ha2018world}. Based on its great functionality and promise, the world model has been used in many aspects such as autonomous driving \cite{ding2024understanding, bar2025navigation, mendonca2023structured, yang2023unisim, guan2024world}, video generation \cite{kang2024far, liu2024sora, mendonca2023structured} and medical\cite{yang2025medical, robertshaw2025world, yue2025echoworld}. 
Existing world models are either continuous action-based \cite{hafner2019dream, saanum2024simplifying, okada2022dreamingv2, hafner2019dream, yang2025instadrive}, relying on a continuously available external control signal to guide and correct state transitions,
or observation-driven \cite{gao2025adaworld, pan2022iso}, learning the dynamics from densely sampled video sequences where the observations themselves serve as a continuous surrogate.
However, continuous interaction and dense observations are costly, or even infeasible in MRI contrast agent kinetics world model, limiting the practicality of these approaches.

\textbf{Spatiotemporal Consistency} has gradually attracted growing attention which can be categorized into two families: Slow feature analysis \cite{wiskott2002slow, jayaraman2016slow, zhang2012slow} assumed that changes between adjacent frames in natural videos are slow and smooth, extracting temporally consistent representations by minimizing temporal derivatives; 
Contrastive learning \cite{qian2021spatiotemporal, Dave_2023_CVPR, tang2023spatio, yuan2022contextualized, ding2022dual, he2025vector, he2025homeomorphism, he2023geometric} is enabled to learn features insensitive to spatiotemporal perturbations through positive-negative sample contrastive constraints.
While both paradigms successfully capture temporal stability in dense video data, they rely on continuous frame supervision and aim at representation robustness rather than generation. Given low temporal resolution training data, the former relies on continuous sampling and thus fails to estimate temporal variation trends, while the latter lacks sufficient samples for comparison, resulting in underfitting of the learned spatiotemporal consistent features.

%% file: Method.tex
\section{Method}
\begin{figure*}[tbp]
    \centering 
    \includegraphics[width=0.97\linewidth]{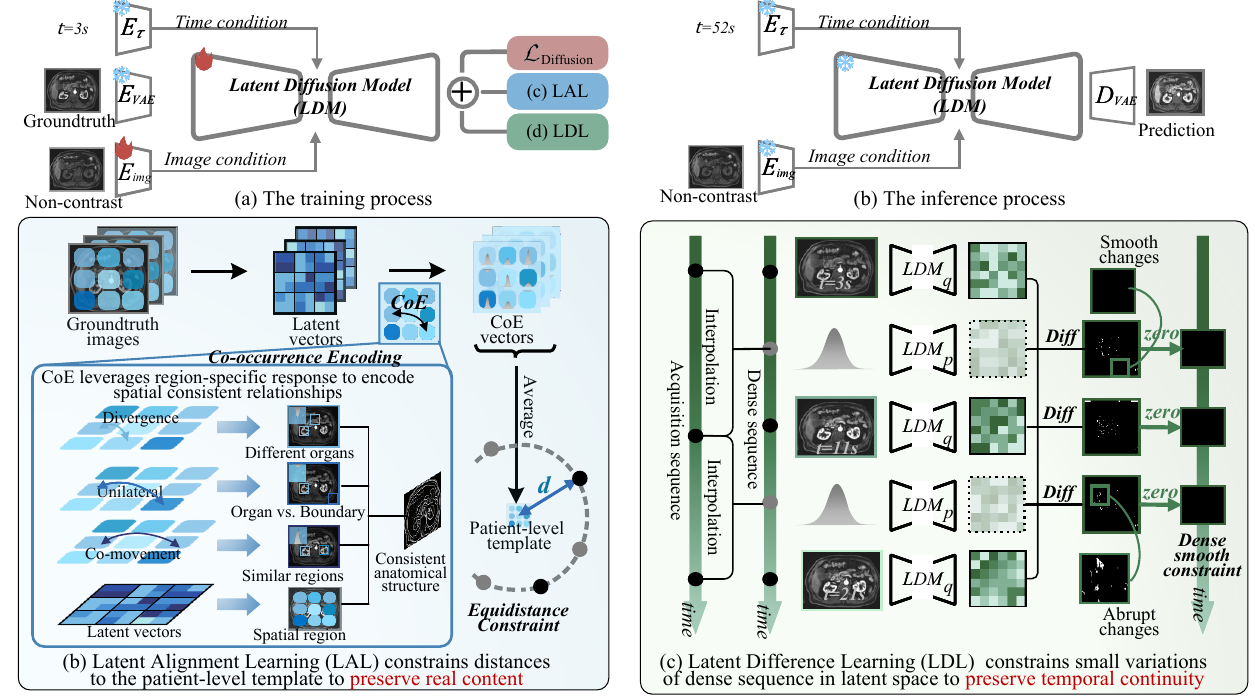} 
    \caption{Overview framework of the MRI CEKWorld. (a) and (b) shows the training and inference processes. (c) LAL captures region-wise co-occurrence relationships and enforces anatomical consistency by aligning to a patient-level template.
    (d) LDL constructs a dense time series in the latent space and imposes a second-order difference (denoted as Diff) on adjacent moments for smooth evolution ($p$ and $q$ denote the inference ).} 
    \label{fig:method} 
\end{figure*}

As shown in Fig.\ref{fig:method}, our spatiotemporal consistency learning implements MRI contrast agent kinetics world model (formulated in Sec.\ref{sec:formulation}) via constraining spatial information at each time points to the patient-level template to preserve content reality (LAL, see Sec.\ref{sec:LAL}) and constraining the latent representation in the dense interpolated sequence to be smooth (LDL, see Sec.\ref{sec:GIL}).

\subsection{Formulation}\label{sec:formulation}
The MRI contrast agent kinetics world model is formulated as an image time series modeling, which predicts the contrast enhanced MRI image $\mathcal{I}(t)$ at arbitrary time $t$ based on a non-contrast image $\mathcal{I}_{p,0}$. 

\textbf{Dataset}
Due to low temporal resolution in MRI acquisition, the dataset is temporally sparsely sampled. 
For each patient $p$, we denote the image-time pairs as
\(\mathcal{D}_p = \{ (\mathcal{I}_{p,i}, t_{p,i}) \}_{i=0}^{T_p}\)
where $\mathcal{I}_{p,i}$ represents the image acquired at time $t_{p,i}$, $T_p$ represents the total of the acquisition time points. 
The complete dataset is defined as 
\(\mathcal{D} = \{ \mathcal{D}_p \mid p = 1, 2, \dots, P \}\) where $P$ is the patient number. 

\textbf{Training}
The model aims to learn a mapping from the initial non-contrast image $\mathcal{I}_{p,0}$ and a continuous time variable $t$ to the corresponding contrast-enhanced image $\mathcal{I}_p(t)$. 
As shown in Fig.\ref{fig:method} (a), $t$, $\mathcal{I}_{p,0}$ and $\mathcal{I}_p(t)$ are encoded separately.
The groundtruth encoder ${E}_{\text{gt}}$,  same as the encoder in VAE \cite{rombach2022high, kingma2013auto} encodes the contrast-enhanced image $\mathcal{I}_p(t)_{\text{gt}}$. 
The time condition encoder ${E}_t$ uses CLIP \cite{radford2021learning} processes the time variable which means the duration after contrast agent injection, converting temporal text information into high-dimensional features that guide the model to generate time-specific enhancement features.  
The image condition encoder ${E}_{img}$, encodes the non-contrast image $\mathcal{I}_{p,0}$ through zero-convolution \cite{zhang2023adding} and adds to the layers in latent diffusion model \cite{rombach2022high}, acting as a hint to guide the prediction.  

Spatial, temporal and diffusion losses are utilized to regularize the generations. The former two will be introduced respectively in Sec.\ref{sec:LAL} and Sec.\ref{sec:GIL}.
The diffusion loss \(
\mathcal{L}_\text{Diffusion} = \mathbb{E}_{t, t, \epsilon} \left[ \left\|\epsilon - \epsilon_\theta\right\|^2 \right] 
\) is used to constrain the accuracy of noise prediction. Here, $t$ represents the denoising timestep.
Let $\mathcal{M}_\theta$ denote the MRI contrast enhancement world model parameterized by $\theta$. 
\begin{equation}
\hat{\mathcal{I}}_p(t) = \mathcal{M}_\theta(\mathcal{I}_{p,0}, t), \quad t \in \mathbb{R}^{+}
\end{equation}

\textbf{Inference}
After training, the optimized model $\mathcal{M}_{\theta^*}$ takes the non-contrast image $\mathcal{I}_{p,0}$ and time $t$  as input to predict the contrast enhanced image:
\(\hat{\mathcal{I}}_p(t) = \mathcal{M}_{\theta^*}(\mathcal{I}_{p,0}, t)\). In Fig.\ref{fig:method} (b), the prediction is decoded by $\mathcal{D}_{img}$ after the U-Net to convert the latent variable into the pixel space.

\subsection{Latent Alignment Learning \textit{\textbf{for real contents}}}\label{sec:LAL}
As shown in Fig.\ref{fig:method} (c), on the basis of anatomical consistency, leveraging the differences in the response patterns of contrast agent signals across various regions, we encode the fluctuation relationship between these regions as a numerical statistical template for anatomical regions, and use this template to constrain the generated results to comply with this fluctuation relationship, suppressing the distortion.

\textit{Co-occurrence Encoding} is implemented by a covariance matrix, which computes the spatial co-occurrence patterns of anatomical structures. Co-movement within regions corresponds to similar regions, while unilateral divergence between regions corresponds to boundary separation. This characterizes the consistent spatial content of the patient. 
Latent representation \(\hat{x}_0\) is extracted by leveraging the reverse process of diffusion models. It uses the model-predicted noise \(\boldsymbol{\epsilon}\) and the noisy sample\(x_t\) to provide a high-quality, structured latent space representation for subsequent statistical calculations and constraints. $ \bar{\alpha}_t = \prod_{s=1}^t \alpha_s $ represents the cumulative coefficient of $ \alpha $ across $ t $ diffusion steps, $x_t$ denotes the noisy sample after $ \tau $ denoising steps:
$\hat{x}_0 = \frac{x_\tau - \sqrt{1 - \bar{\alpha}_\tau} \cdot \boldsymbol{\epsilon}}{\sqrt{\bar{\alpha}_\tau}}$.
Then, in the latent space, we have the prediction series $\hat{x_0} = \{\hat{x_0}_t \in \mathbb{R}^{c \times h \times w}\}_{t=1}^T$, where $c$ is the number of channels, $h$ and $w$ are the height and width of $\hat{x}_0$.
We flatten each time point of $\hat{x_0}_t$ into $X_t \in \mathbb{R}^{c \times s}$ with $s = h \cdot w$ and center it along the spatial dimension
$X_t^c = X_t - \frac{1}{s} \sum_{s_i=1}^s X_t$.
This removes the spatial mean bias so that the covariance reflects the true distribution shape of the features.
Covariance matrix for each time acquisition time point $t$ is computed as $\Sigma_t = \frac{1}{S-1} X_t^c (X_t^c)^\top$
and then regularized with shrinkage and a small jitter to ensure positive-definiteness, so
$\tilde{\Sigma}_t = (1-\gamma)\Sigma_t + \gamma I + \varepsilon I$
where $\gamma$ controls the shrinkage strength, $I$ is the identity matrix, and $\varepsilon$ is a small jitter.

The patient-level template, obtained by computing the mean of covariance matrix of time points, represents a more stable patient-level spatial feature under the spatial law of invariant patient anatomical structures. 
We map each latent covariance $\Sigma_t$ to an Euclidean vector via the log–Cholesky parameterization for numerical stability and positive–definiteness preservation for optimization in training \cite{lin2019riemannian}. 
Let $L_t=\mathrm{chol}(\Sigma_t)$; extract $\mathrm{lower}_t=\mathrm{vec}(\mathrm{tril}(L_t,-1))$ and $\mathrm{logdiag}_t=\log(\mathrm{diag}(L_t))$, and form $z_t=[\,\mathrm{lower}_t;\,\mathrm{logdiag}_t\,]$. Averaging gives the patient-level template vector $\bar z=\frac{1}{T}\sum_{t=1}^{T} z_t$, which we use subsequently as the template representation.

\textit{Equidistance constraint} constrains the $z_t$ at different time points are consistent with the template by keeping the same distance from the template $\bar z$. It not only ensures statistical consistency but also allows reasonable dynamic changes to be preserved between time points, so that the generated sequence has a real spatial content under sparse supervision.
The spatial loss from equidistance constraint of distance \(d_t^2 = \big\| z_t - \bar z \big\|_2^2\) is defined as:
\begin{equation}
\mathcal{L}_{\text{Spatial}}
\;=\;
\frac{1}{P}\sum_{p=1}^{P}\;
\frac{1}{T_p}\sum_{t=1}^{T_p}
\Big(d_{p,t}^{\,2}\Big).
\label{eq:Lspatial_equidistance_patientwise}
\end{equation}
where $P$ represents the total patient number and $T_p$ represents the total acquisition time points of the $p$-th patient.

\subsection{Latent Difference Learning \textbf{\textit{for continuity}}} \label{sec:GIL}
As shown in Fig.\ref{fig:method}(d), LDL generates predictions in the latent space for the unobserved intermediate time points by interpolation, and imposes smoothness constraints on the interpolated dense sequence for semantic continuity to achieve temporal smooth transition.

Intermediate points are inserted to generate a dense sequence in the latent space.
The set of original sparse acquisition time points consists of real observed time values, defined as:
$T_{\text{acq}} = \{ t_{\text{acq}, 0}, t_{\text{acq}, 1}, \dots, t_{\text{acq}, N-1} \}$
where \( N \) is the number of original acquisition time points; \( t_{\text{acq}, i}\) denotes the absolute time of the \( i \)-th original observation  and ordered by acquisition time.
For each pair of adjacent original observations \( (t_{\text{acq}, i}, t_{\text{ac1}, i+1}) \) (\( i \in [0, N-2] \)), let \( K_i \) be the number of inserted intermediate points (\( K_i \geq 1 \)),  on the time interval \( \Delta t_i = t_{\text{acq}, i+1} - t_{\text{acq}, i} \), uniform interpolation to ensure intermediate points lie strictly between \( t_{\text{acq}, i} \) and \( t_{\text{acq}, i+1} \). The time value of the \( k \)-th intermediate point (\( k \in [1, K_i] \)) between the \( i \)-th pair of adjacent points is:
$t_{\text{mid}, i, k} = t_{\text{acq}, i} + \frac{k}{K_i + 1} \cdot \Delta t_i$.
The dense sequence \( T_{\text{dense}} \) is the union of original observations and all intermediate points, sorted by time:
$T_{\text{dense}} = T_{\text{acq}} \cup \left( \bigcup_{i=0}^{N-2} \{ t_{\text{mid}, i, 1}, \dots, t_{\text{mid}, i, K_i} \} \right)$.

Dense prediction sequences in the latent space are constructed as a set of clean predictions at all dense time points, including both the
outputs anchored at real acquisition times and those generated from noise at inserted intermediate times. 
For each observed acquisition point \(t_{\text{acq},i}\), we recover its latent vector \(\hat{x}_{0,\text{acq},i}\) from the corresponding noisy sample \(x_{\tau,\text{acq},i}\) after $\tau$ denoising steps:
\(
\hat{x}_{0,\text{acq},i}
= \frac{x_{\tau,\text{acq},i}-\sqrt{1-\bar{\alpha}_{\tau_{\text{acq},i}}}\,\epsilon_\theta}{\sqrt{\bar{\alpha}_{\tau_{\text{acq},i}}}} .
\)
For each inserted intermediate time point \(t_{\text{mid},j}\), we sample a noise latent \(x_{\tau,\text{mid},j}\sim\mathcal{N}(0,I)\) at the assigned timestep \(\tau_{\text{dense},j}\), and obtain its latent prediction by applying denoising schedule of the latent diffusion model \cite{zhang2023adding}:
\(
\hat{x}_{0,\text{mid},j}
= _\theta(x_{\tau,\text{mid},j},\, \tau_{\text{dense},j}),
\)
where \(p_\theta\) denotes the standard inference denoising process that maps a noisy latent to its predicted latent vector \cite{zhang2023adding}.
Thus, the dense latent prediction sequence is finally assembled as
\[
\hat{X}_{\text{dense}}[j]=
\begin{cases}
\hat{x}_{0,\text{acq},i}, & t_{\text{dense},j}=t_{\text{acq},i},\\[3pt]
\hat{x}_{0,\text{mid},j}, & t_{\text{dense},j}\text{ is intermediate}.
\end{cases}
\]

\textit{Dense smooth constraint} limits the abrupt variance to zero through second-order difference center difference for smoothness over time. After de-duplication to remove possible duplicate time points of \( T_{\text{dense}}\), we obtain an ordered time sequence \( T_{\text{sort}} = \{ t_0, t_1, \dots, t_{T-1} \} \) (\( T \leq M \), where \( t_k \) is the time value in seconds) and corresponding model outputs \( \boldsymbol{y}_{\text{sort}} = \{ y_{\text{sort}}^{0}, y_{\text{sort}}^{1}, \dots, y_{\text{sort}}^{T-1} \} \) (where \( y_{\text{sort}}^{k} \in \mathbb{R}^{1 \times c \times h \times w} \) is the model output at \( t_k \)). For each point \( k \in [1, T-2] \) , the dense smooth constraint among discrete different time points through the center difference equation is derived in \textit{Supplementary},  which is defined as:
\begin{equation}
\begin{split}
\textbf{D}_2^{k} &= 2 \cdot \bigg( \frac{y_{\text{sort}}^{k-1}}{h_0^{k} \cdot (h_0^{k} + h_1^{k})}  \quad - \frac{y_{\text{sort}}^{k}}{h_0^{k} \cdot h_1^{k}} \\
& \quad + \frac{y_{\text{sort}}^{k+1}}{h_1^{k} \cdot (h_0^{k} + h_1^{k})} \bigg) \cdot w^{k}
\end{split}
\label{eq:center_difference}
\end{equation}
where \( h_0^{k} = t_k - t_{k-1} + \delta \) and \( h_1^{k} = t_{k+1} - t_k + \delta \) (\( \delta = 10^{-6} \) to avoid division by zero) are adjacent time intervals; \( w^{k} = \frac{1}{1 + h_0^{k} + h_1^{k}} \) is the interval weight used for weaker penalty for larger intervals, adapting to varying temporal densities. The final loss is the average of these differences using the L1 norm for robustness to outliers, which is expressed as 
\begin{equation}
\mathcal{L}_{\text{Temporal}} = \frac{1}{T-2} \sum_{k=1}^{T-2} \| \textbf{D}_2^{(k)} \|_1 .
\label{eq:temporal}
\end{equation}

%% file: Experiments.tex
\section{Experiments}
\label{sec:experiments}
\subsection{Comparison Study}
\subsubsection{Experiment Protocol} This section introduces the overview protocol in our study for complete and fair evaluations in our experiments.

\begin{table*}[t]
\centering
\caption{Quantitative results: Quantitative comparison of different methods on Abdominal and Breast DCE-MRI datasets show that our method achieves the best performance. Quantitative ablation results verify the effectiveness of our LAL and LDL. “Avg.SSIM" and "Avg.cSSIM" denote the average score over two datasets in spatial SSIM and temporal cSSIM metrics, respectively.}
\label{tab:method_comparison_dce_mri}

\resizebox{\textwidth}{!}{
\begin{tabular}{l c c c c c c c c c c  c  c}
\toprule
\multirow{2}{*}{Method} 
& \multicolumn{5}{c}{\textbf{Abdominal DCE-MRI}} 
& \multicolumn{5}{c}{\textbf{Breast DCE-MRI}}
& \multirow{2}{*}{\makecell[c]{\textbf{Avg}.\\SSIM$\uparrow$}}
& \multirow{2}{*}{\makecell[c]{\textbf{Avg}.\\cSSIM$\uparrow$}}
\\
\cmidrule(lr){2-6} \cmidrule(lr){7-11}
& PSNR$\uparrow$ & SSIM$\uparrow$ & LPIPS$\downarrow$ & rMSE$\downarrow$ & cSSIM$\uparrow$
& PSNR$\uparrow$ & SSIM$\uparrow$ & LPIPS$\downarrow$ & rMSE$\downarrow$ & cSSIM$\uparrow$
&  &  \\ 
\midrule

CustomDiff\cite{kumari2023multi}     
& 17.73 & 0.4130 & 0.4646 & 295.7 & 0.3551 
& 11.19 & 0.2463 & 0.4706 & 1094 & 0.3835 
& 0.3296 
& 0.3693 \\

T2I\cite{mou2024t2i}    
& 16.89 & 0.4396 & 0.3100 & 124.1 & 0.3396 
& 17.73 & 0.4130 & 0.2905 & 297.4 & 0.1347 
& 0.4263
& 0.2372 \\

CCNet\cite{osuala2024towards} 
& \textbf{24.35} & 0.5794 & 0.2735 & 43.44 & 0.7098
& \textbf{21.47} & 0.4043 & 0.3128 & 289.3 & 0.3155
& 0.4918
& 0.5127 \\

EditAR\cite{mu2025editar}
& 22.65 & 0.5571 & 0.3314 & 49.76 & 0.7536
& 19.85 & 0.4170 & 0.3119 & 288.9 & 0.3886
& 0.4870
& 0.5711 \\

ControlNet\textit{$_{baseline}$} \cite{zhang2023adding} 
& 23.61 & 0.7178 & 0.2719 & 43.08 & 0.8286
& 19.79 & 0.5196 & 0.2625 & 308.0 & 0.3370
& 0.6187
& 0.5828 \\

\midrule

+ \textit{LAL} 
& 23.92 & 0.7227 & 0.2666 & 40.29 & 0.8439
& 20.86 & 0.5442 & 0.2640 & 250.7 & 0.3879
& 0.6335
& 0.6159 \\

+ \textit{LDL} 
& 24.05 & 0.7369 & 0.2623 & 40.25 & 0.8411
& 20.21 & 0.5391 & 0.2636 & 261.5 & 0.3392
& 0.6380
& 0.5901 \\

\textbf{MRI CEKWorld}
& 24.06 & \textbf{0.7419} & \textbf{0.2622} & \textbf{40.08} & \textbf{0.8451}
& 21.09 & \textbf{0.5599} & \textbf{0.2620} & \textbf{243.5} & \textbf{0.3900}
& \textbf{0.6509}
& \textbf{0.6176} \\

\bottomrule
\end{tabular}
}
\end{table*}

\textbf{Datasets}  Two Dynamic Contrast Enhancement - Magnetic Resonance Imaging (DCE-MRI) datasets are used: (1) Private Abdominal DCE-MRI Dataset (Abdominal DCE-MRI): This abdominal consists of 91 patients. There is one non-contrast image, and 15 contrast enhanced images within 300 seconds after contrast agent injection. Among these contrast enhanced images, 6 are in the arterial phase, 6 are in the venous phase, and 3 are in the delayed phase.
(2) Public Duke Breast DCE-MRI Dataset (Breast DCE-MRI) \cite{saha2021dynamic}: This dataset contains 922 examination records of breast DCE-MRI. After the injection of contrast agent, contrast-enhanced data at 3 or 4 time points are acquired. Following \cite{osuala2024towards}, we crop the slices containing the lesion region and increase the width and height of the tumor bounding box to half the width and height of the full image.
Both datasets are resized to the 256$\times$256, normalized to $[-1,1]$ \cite{rombach2022high} and then stacked into 3 channels as the image input.
Since the acquisition time of DCE-MRI sequences in both datasets is manually controlled, the acquisition of sequences is not strictly fixed at specific time points. Thus, the time points in the test set rarely appear in the training set.

\textbf{Evaluation Metrics}
Both spatial and temporal metrics are utilized to validate the performance of virtual MRI contrast enhancement prediction in all experiments.
\textit{Spatially},  following the studies \cite{zhang2023adding, danier2024ldmvfi, osuala2024towards, kim2024data}, Peak Signal-to-Noise Ratio (PSNR), Structural SIMilarity (SSIM), Learned Perceptual Image Patch Similarity (LPIPS) and root Mean Squared Error (rMSE) are used to evaluate in all experiments. 
\textit{Temporally}, continuous SSIM (cSSIM) is designed to quantify the structural consistency between adjacent frames in the temporal dimension. Assuming a time series contains \( N \) consecutive frames , corresponding to time points \( t=1,2,...,N \), denoted as \( I_1, I_2, ..., I_N \) with adjacent frame pairs \( (I_t, I_{t+1}) \) (a total of \( N-1 \) pairs), its formula is:  
\(cSSIM = \frac{1}{N-1} \sum_{t=1}^{N-1} \text{SSIM}(I_t, I_{t+1})
 \)
where \( \text{SSIM}(I_t, I_{t+1}) \) is the structural similarity of a single pair of adjacent frames. 
We then define a spatial average score Avg.SSIM by averaging the SSIM over both datasets, and a temporal average score Avg.cSSIM by averaging the cSSIM over both datasets. Both average metrics lie in [0,1], with larger values indicating better overall performance.

\textbf{Implementation Details}
We adopt the ControlNet \cite{zhang2023adding} as the model backbone and make U-Net \cite{rombach2022high}, image encoder ${E}_{img}$ trainable as well. Training and testing were implemented on an NVIDIA A100 GPU with 40GB of memory. For all of the hyper parameters in training process, total epoch is 14, batch size is 4, $\lambda_{\text{Spatial}}=6.0$, $\lambda_{\text{Temporal}}=1.0$, the $K_i=2$ for the best performance in Abdominal DCE-MRI;  $\lambda_{\text{Spatial}}=4.0$, $\lambda_{\text{Temporal}}=1.0$, the $K_i=2$ obtains the best performance in Breast DCE-MRI. The whole pipeline is two-stage,
first stage performs a diffusion warm-up to stabilize the initial latent space, then introduces spatial regularization to establish patient-specific content. The loss function is \(
\mathcal{L}_{1}
= \mathcal{L}_{\mathrm{Diffusion}} + \lambda_{\mathrm{Spatial}}\mathcal{L}_{\text{Spatial}}\)
where $\lambda_{\text{Spatial}}$ represents the hyperparameter for the strength of spatial regularization. The second stage switches to temporal regularization to leverage the aligned content and achieve smooth transitions.
The loss of the second stage is \(
\mathcal{L}_{2}
= \mathcal{L}_{\mathrm{Diffusion}} + \lambda_{\mathrm{Temporal}}\mathcal{L}_{\text{Temporal}}\).

\textbf{Comparisons Setting}
\begin{figure*}[tbp]
    \centering 
    \includegraphics[width=0.97\linewidth]{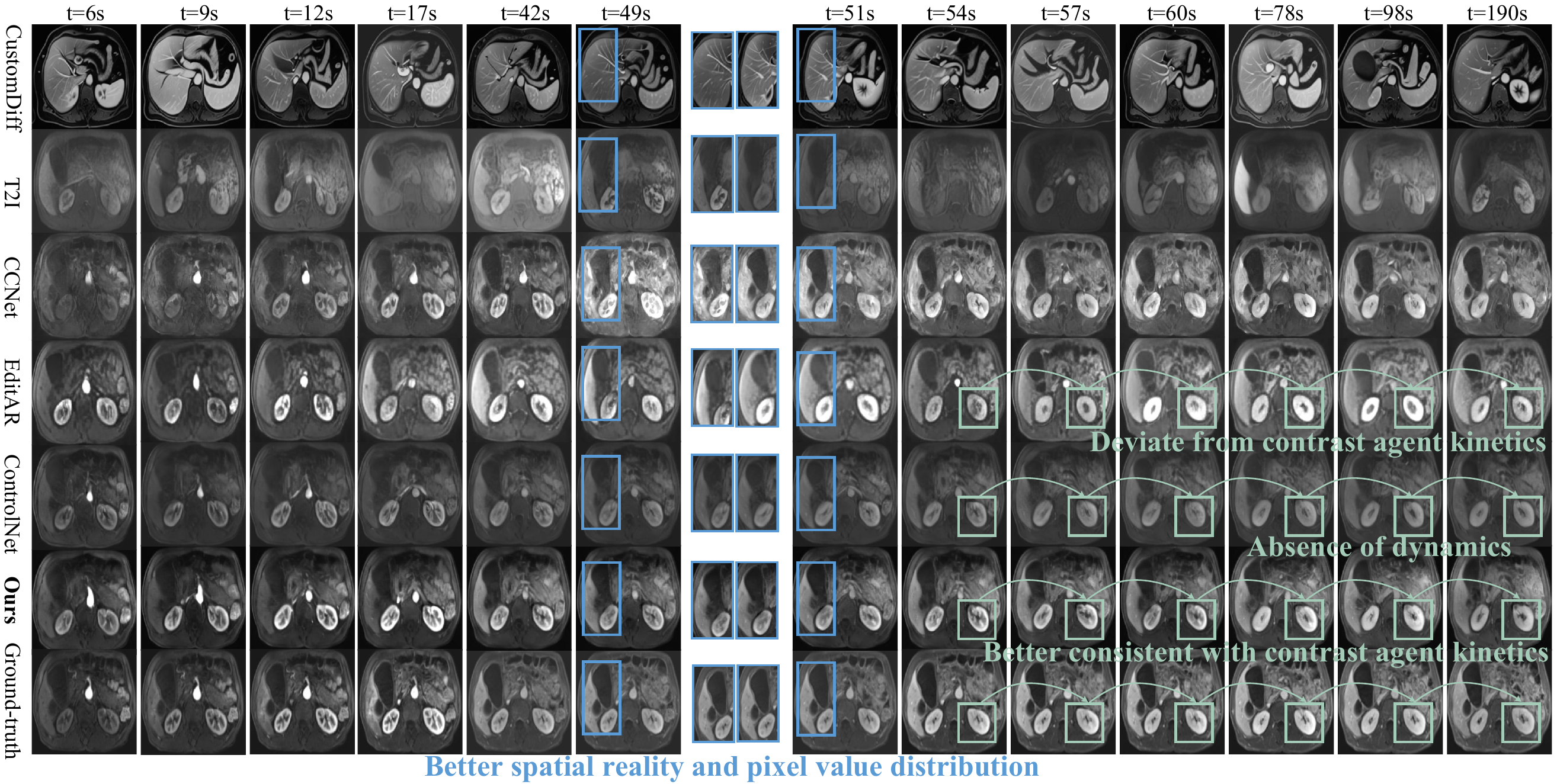}
    \caption{\textbf{Visualization results} in Abdominal DCE-MRI: The visualization results of our methods on different time points exhibit better spatial reality (zoom-in regions in blue boxes) and temporal continuities than comparisons (connected green boxes), whereas other methods show deviations from realistic kinetics or lack of dynamic consistency.} 
    \label{fig:compare} 
\end{figure*}
Recent controllable image generation frameworks \cite{kumari2023multi, mou2024t2i, zhang2023adding, mu2025editar}. DCE-MRI contrast enhancement method \cite{osuala2024towards}, ContrastControlNet (CCNet) are compared.  All the hyperparameter experiment settings are the same, the time inputs are formulated as "HH:MM:SS", where HH, MM, SS mean the hours, minutes and seconds of the time interval between the pre-contrast images and the generation. CCNet, T2I and ControlNet are all based on SD1.5 \cite{rombach2022high} following \cite{mu2025editar}.

\subsubsection{Comparison Analysis}
\textbf{Quantitative Results Analysis}
Quantitative results on both datasets show our method outperforms others in spatial fidelity and temporal smoothness (Tab.~\ref{tab:method_comparison_dce_mri}), leading in Avg.S and Avg.T.
\textit{Spatially}, our LAL module preserves anatomical consistency, achieving the best SSIM, LPIPS, and rMSE. While CCNet attains high PSNR, it fails to fully converge under the same training settings—producing over-smoothed predictions that lose structural detail, hence poor SSIM, LPIPS, and rMSE. Excluding this case, our method reaches 24.06 PSNR, 0.7419 SSIM, and 0.2622 LPIPS on the Abdominal DCE-MRI dataset, and leads all spatial metrics on the Breast dataset. The higher rMSE of the Breast dataset (intensity range ~0–4000) stems from data distribution rather than generation performance.
\textit{Temporally}, our method achieves the highest cSSIM (0.8451 for Abdominal, 0.3900 for Breast DCE-MRI), preserving inter-frame structural coherence with smoother enhancement kinetics and fewer abrupt intensity changes.

\begin{figure}[tbp]
    \centering
    \includegraphics[width=0.97\linewidth]{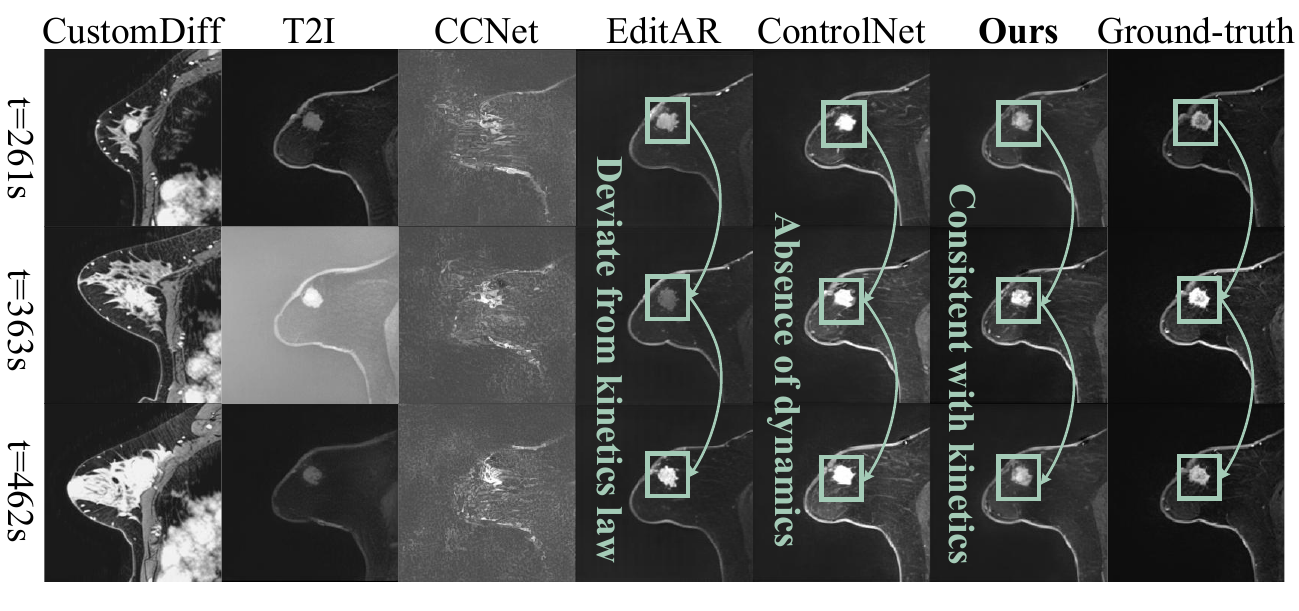} 
    \caption{\textbf{Visualization results} in Breast DCE-MRI: The visualization results achieves temporal consistent (connected green boxes) with contrast agent kinetics, demonstrating superior fidelity in breast DCE-MRI sequence generation.}
    \label{fig:duke_compare} 
\end{figure}

\textbf{Visualization Results Analysis}
As shown in Fig.\ref{fig:compare} and Fig.\ref{fig:duke_compare}, the visualization sequences of both datasets demonstrate that our method achieves high spatial reality and natural kinetics, both closely matching the ground-truth.
CustomDiff and T2I generated images with severe deviations from the ground-truth, suffering from blurred organ contours and distorted dynamic enhancement gradients of contrast agents.
CCNet failed to converge, leading to excessively smooth images, severe spatial structural distortion, and the appearance of color blocks, which is consistent with the previously analyzed characteristic of high PSNR. EditAR and ControlNet have normal spatial structures, but their kinetics both deviate from the normal pattern. More visualization results are in \textit{Supplementary}.

\subsection{Ablation Study and Model Analysis}
\textbf{Component Ablation}
The ablation studies in both datasets show the effectiveness of our proposed innovations. As shown in the downside part of Table.\ref{tab:method_comparison_dce_mri}, the LAL achieves 2.46\% SSIM and 1.07 PSNR improvement in Breast DCE-MRI, which demonstrates the effectiveness of consistency owing to LAL. If using alone, significant promotion also represents its temporal smoothness compared with baseline, which achieves 1.25\% SSIM in Abdominal DCE-MRI and 5.09\% improvement in Breast DCE-MRI.
When combining two innovations, the improvement results show that under the premise that LAL has formed better spatial structural consistency, further enhances both temporal dynamic smoothness and spatial structural consistency.

\begin{figure}[tp]
    \centering
    \includegraphics[width=0.97\linewidth]{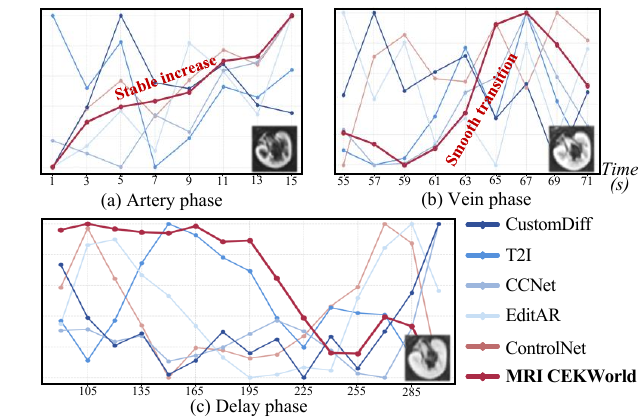} 
    \caption{\textbf{Contrast agent kinetics time curve}: The curves in (a) artery phase, (b) vein phase and (c) delay phase are more stable and smooth than compared methods.}
    \label{fig:smooth} 
\end{figure}

\textbf{Contrast Agent Kinetics Time Curve} As shown in Fig.\ref{fig:smooth}, the performance of the curves across the three key phases in clinical use demonstrates out method has a stronger capability in modeling the contrast agent kinetics.
We performed equidistant sampling for the artery phase (1–15 s), vein phase (55–72 s), and delay phase (90–300 s) according to the common acquisition time. The sampling results of the mean gray value at each time point for the renal region of interest from each method were normalized to compare their smoothness and stability.
In (a) artery phase, MRI CEKWorld exhibits a stable increase, which precisely matches the physiological process of rapid contrast agent filling in the artery phase. 
In (b) vein phase, ours shows a curve pattern of smooth transition, which reflects its accurate capture of the kinetics process of contrast agent in the vein phase. The curves increase by the accumulation of contrast agent within the interstitial space of the renal parenchyma then decrease owing to the washout phase. In contrast, competing methods such as CCNet and EditAR exhibit obvious abrupt fluctuations in their curves.  
In (c) delay phase, ours first maintain a stable signal level and then decays smoothly over time because this is fully consistent with the physiological mechanism of gradual contrast agent clearance in the delay phase, verifying its reliable modeling capability for long-term enhancement dynamics. 

\begin{figure}[tbp]
    \centering
    \includegraphics[width=0.97\linewidth]{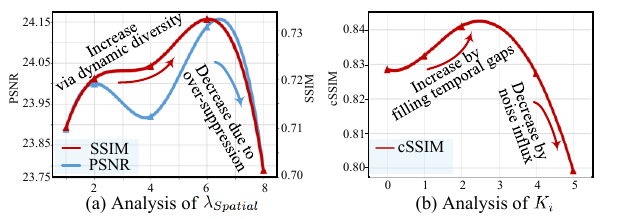} 
    \caption{\textbf{Hyperparameter analysis}: Analysis of $\lambda_{Spatial}$ and $K_i$ shows a trend of first increasing and then decreasing. (a) PSNR and SSIM increase by allowing more dynamic diversity, decrease due to over-suppression on the dynamic. (b) cSSIM increases by filling temporal gaps, and decreases due to noise influx.}
    \label{fig:hyper}
\end{figure}

\textbf{Hyper-parameter Ablation of $\lambda_{\text{Spatial}}$} In Fig.\ref{fig:hyper} (a), the transition of \(\lambda_{\text{Spatial}}\) shows a trend of first increasing and then decreasing in terms of SSIM and PSNR metrics. Since \(\lambda_{\text{Spatial}}\) determines the spatial regularization strength of LAL, when \(\lambda_{\text{spatial}}\) is small, the constraint of distance consistency is weak, and the generated results deviate from the template; when \(\lambda_{\text{Spatial}}\) is moderate which $\lambda_{\text{Spatial}}$ is 6, the constraint strength achieves a balance between following the template and preserving feature diversity; when \(\lambda_{\text{Spatial}}\) is excessively large, this constraint rigidly enforces features to stay close to the template, suppressing the reasonable feature differences that should exist between time points.

\textbf{Hyper-parameter Ablation of $K_i$} As shown in Fig.\ref{fig:hyper}, the variation of \(K_i\) also shows a first increasing then decreasing trend in terms of cSSIM that measures continuity. With the increase of \(K_i\), the newly added intermediate sampling points exactly fill the gaps in the sparse temporal sequence, which provides the model with more refined intermediate states in temporal evolution, enabling it to more accurately learn the continuous changes of contrast agent kinetic laws. However, when \(K_i\) exceeds 2, an excessive number of intermediate sampling points do not come from the real data distribution and carry noise that deviates from real patterns. This interferes with the model's learning of real temporal features, leading to a subsequent decrease.

\begin{figure}[htbp]
    \centering
    \includegraphics[width=0.47\textwidth]{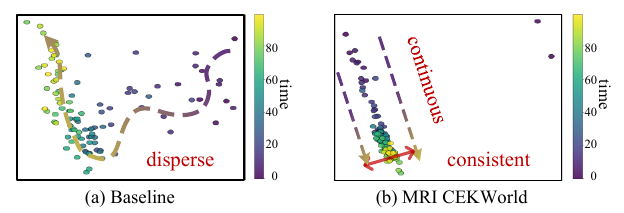} 
    \caption{\textbf{Distribution in latent space}: The distribution of feature points in the latent space demonstrates generations of continuous time points adhere to temporal continuity and spatial consistency.}
    \label{fig:latent} 
\end{figure}

\textbf{Latent Representation at Continuous Time Points} As shown in Fig.\ref{fig:latent}, the distribution of continuous sequences generated by MRI CEKWorld is continuous and consistent, which demonstrates the spatial consistency and temporal continuity are preserved in the latent space. We visualize such a distribution by compressing the latent space vectors obtained from the latent space in ControlNet during the reverse process into a low-dimensional space via principal component analysis, and using the corresponding time for coloring (the lighter and yellower the color, the larger the time value). In Fig.\ref{fig:latent} (a), the dots show a disperse state, indicating that their latent features have no obvious temporal pattern in the low-dimensional space and the feature distribution across different time points is chaotic. In contrast, our dots are distributed consistently and continuously, which suggests that the features at different time points have a strong clustering property, indicating that the consistency is stably preserved over time. Furthermore, as time progresses, the color sequentially changes from light to dark, reflecting that the features transition smoothly. It is worth noting that the outliers in the upper right corner of Fig.\ref{fig:latent} (b) correspond to the feature points at \( = 0 \, \text{s}\) and \( = 1 \, \text{s}\). Due to the limitation of the central difference in Eq.\ref{eq:center_difference}, the constraints on these two points are neglected.

%% file: Conclusion.tex
\section{Conclusion}
In this paper, we introduced MRI CEKWorld, the first contrast enhancement kinetics world model designed to simulate contrast agent kinetics in human body for the inefficient information yield in clinical MRI acquisition. Exploiting the inherent spatiotemporal consistency of contrast enhancement, we devised a spatiotemporal consistency learning under a sparsely sampled dataset which includes the latent alignment and difference learning. Despite its strong performance, we will extend to other contrast-enhanced imaging modalities, such as computed tomography, aiming for a unified contrast kinetics world model.
\label{sec:conclusion}

%% file: X_suppl.tex
\clearpage
\setcounter{page}{1}
\maketitlesupplementary

\section{The Derivation of Dense Smooth Constraint}
We regularize the second-order temporal difference of the output sequence to suppress sudden jumps without excessively weakening real contrast changes. The first-order difference directly penalizes the magnitude of changes, making the model tend to generate nearly static sequences. In contrast, the second-order difference describes the rate of change of velocity, allowing for monotonic enhancement/attenuation when such changes are smooth. 
Since the sampling time points in this task are non-uniform, we further derive the corresponding three-point central difference formula on a non-uniform time grid, as shown below.

Let $\{\tau_{k-1},\tau_k,\tau_{k+1}\}$ be three consecutive time stamps and denote adjacent gaps by $h_0^k=\tau_k-\tau_{k-1}$ and $h_1^k=\tau_{k+1}-\tau_k$. Expanding $y(\tau_{k-1})$ and $y(\tau_{k+1})$ at $\tau_k$ and keeping the remainder terms,
\begin{align*}
y(\tau_{k-1})&=y(\tau_k)-h_0^k y'(\tau_k)+\tfrac{(h_0^k)^2}{2}y''(\tau_k)-\tfrac{(h_0^k)^3}{6}y'''(\xi_0),\notag\\
y(\tau_{k+1})&=y(\tau_k)+h_1^k y'(\tau_k)+\tfrac{(h_1^k)^2}{2}y''(\tau_k)+\tfrac{(h_1^k)^3}{6}y'''(\xi_1),
\end{align*}
for some $\xi_0\!\in\!(\tau_{k-1},\tau_k)$ and $\xi_1\!\in\!(\tau_k,\tau_{k+1})$. Multiplying the first expansion by $h_1^k$ and the second by $h_0^k$ cancels $y'(\tau_k)$; neglecting $O\!\big((h_0^k)^3+(h_1^k)^3\big)$ then yields the nonuniform three-point central difference
\begin{align*}
y''(\tau_k)\approx {}&
\frac{2}{h_0^k h_1^k (h_0^k+h_1^k)} \\&
\cdot \Big[h_1^k\!\big(y(\tau_k)-y(\tau_{k-1})\big)
   +h_0^k\!\big(y(\tau_{k+1})-y(\tau_k)\big)\Big]
\notag\\
&=2 \left(
\frac{y(\tau_{k-1})}{h_0^k(h_0^k+h_1^k)}
\right.
\notag\\
&\quad\left.
-\frac{y(\tau_k)}{h_0^k h_1^k}
+\frac{y(\tau_{k+1})}{h_1^k(h_0^k+h_1^k)}
\right).
\end{align*}

Notation by setting $y(\tau_j)=y_{\text{sort}}^{\,j}$ and adding a small stabilization $\delta$ to avoid zero denominators,
\begin{equation}
h_0^k=\tau_k-\tau_{k-1}+\delta,\qquad
h_1^k=\tau_{k+1}-\tau_k+\delta,
\end{equation}
we weight each stencil by $w^k=\tfrac{1}{1+h_0^k+h_1^k}$ to reduce the penalty over large temporal gaps. The dense smoothness (center-difference) operator used in the loss at index $k$ is therefore
\begin{align*}
\mathbf{D}_2^{k}
= {}& 2\!\left(
\frac{y_{\text{sort}}^{\,k-1}}{h_0^{k}(h_0^{k}+h_1^{k})}
-\frac{y_{\text{sort}}^{\,k}}{h_0^{k}h_1^{k}}
+\frac{y_{\text{sort}}^{\,k+1}}{h_1^{k}(h_0^{k}+h_1^{k})}
\right)\! w^{k},
\end{align*}
which is a second-order accurate approximation of $y''(\tau_k)$ on a nonuniform grid and reduces to the standard centered stencil when $h_0^k=h_1^k$.

\section{The Details of Experiments}
\subsection{Datasets}
\textbf{Abdominal DCE-MRI}
It is acquired via the CAIPIRINHA-Dixon-TWIST-VIBE technique\cite{hao2020image}, which fully covers multiple time phases before and after contrast agent injection to reflect tissues’ hemodynamic characteristics accurately.
Specifically, it comprises 1 pre-contrast phase before contrast agent injection, approximately 6 random arterial phases between 15 and 37 seconds post contrast agent injection for capturing the rapid filling process of contrast agent in the arterial system, approximately 6 random portal vein phases between 50 and 72 seconds post contrast agent injection for reflecting the distribution and perfusion of contrast agent into the portal vein system, and several delayed phases which are acquired nearby 90 s, 150 s, or 300 s post contrast agent injection, respectively, to observe the delayed enhancement or clearance of contrast agent in tissues. 
Since the raw data were not registered, all enhanced sequences in the dataset were registered to the non-contrast images.
\textbf{Breast DCE-MRI}
This dataset involves 2 manufacturers. Among them, equipment from GE Medical Systems accounts for a higher proportion, equipped with two magnetic field strengths: 1.5T and 3T. Equipment from Siemens mainly includes models such as Avanto, Skyra, and TrioTim, which also cover both 1.5T and 3T magnetic field strengths. A median of 131s passed between DCE sequences  \cite{saha2021dynamic}. 

\begin{figure*}[htbp]
    \centering 
    \includegraphics[width=1.0\linewidth]{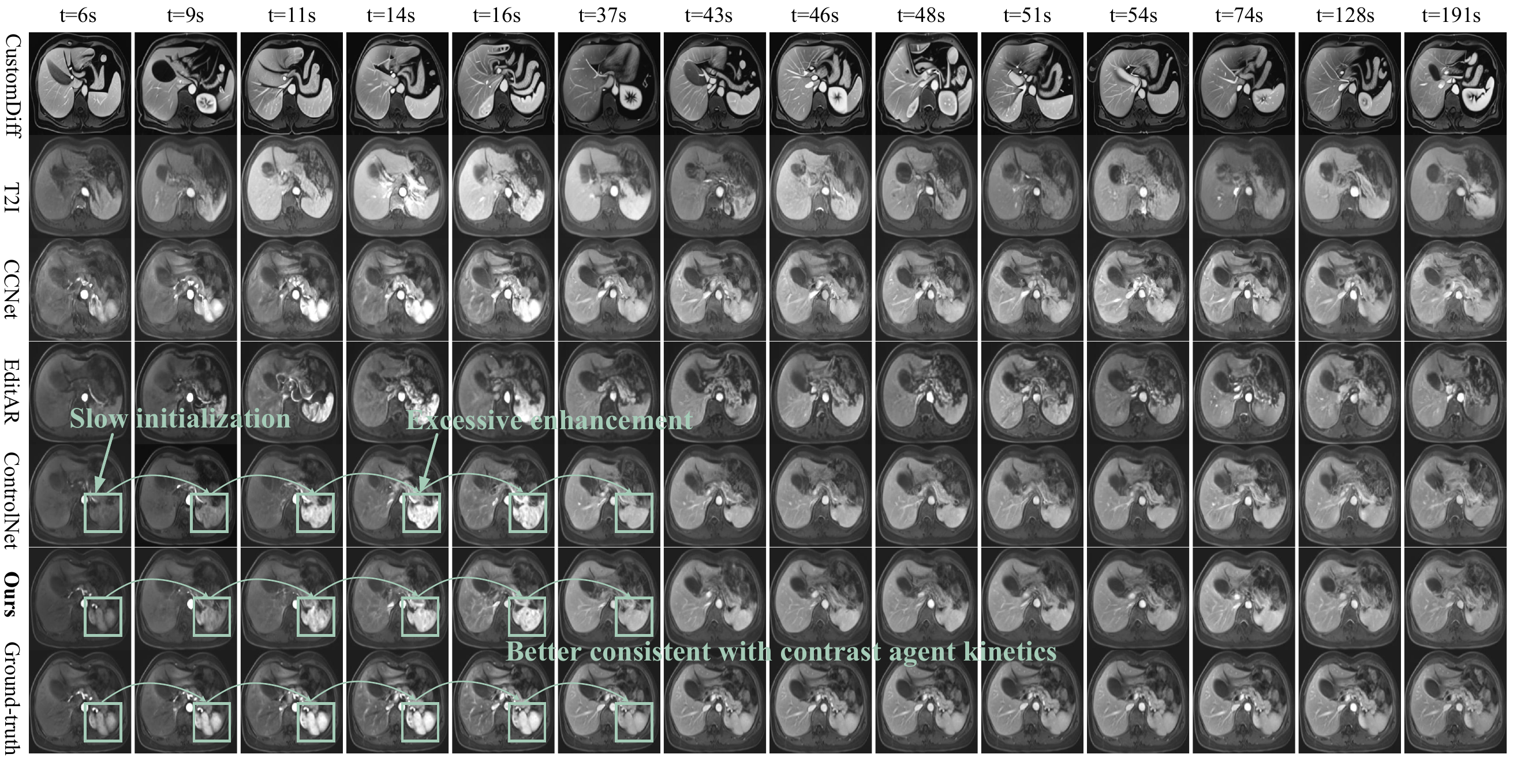}
    \caption{\textbf{Visualization results of spleen} in Abdominal DCE-MRI shows that our CEKWorld conforms to the spleen contrast agent kinetics, exhibiting the characteristics of rapid synchronous enhancement, progressive homogenization, and sustained homogeneous high signal. In contrast, methods such as ControlNet and CCNet have defects, including delayed start of enhancement, abnormal intensity, or content distortion.} 
    \label{fig:compare_spleen} 
\end{figure*}

\begin{figure*}[h]
    \centering 
    \includegraphics[width=1.0\linewidth]{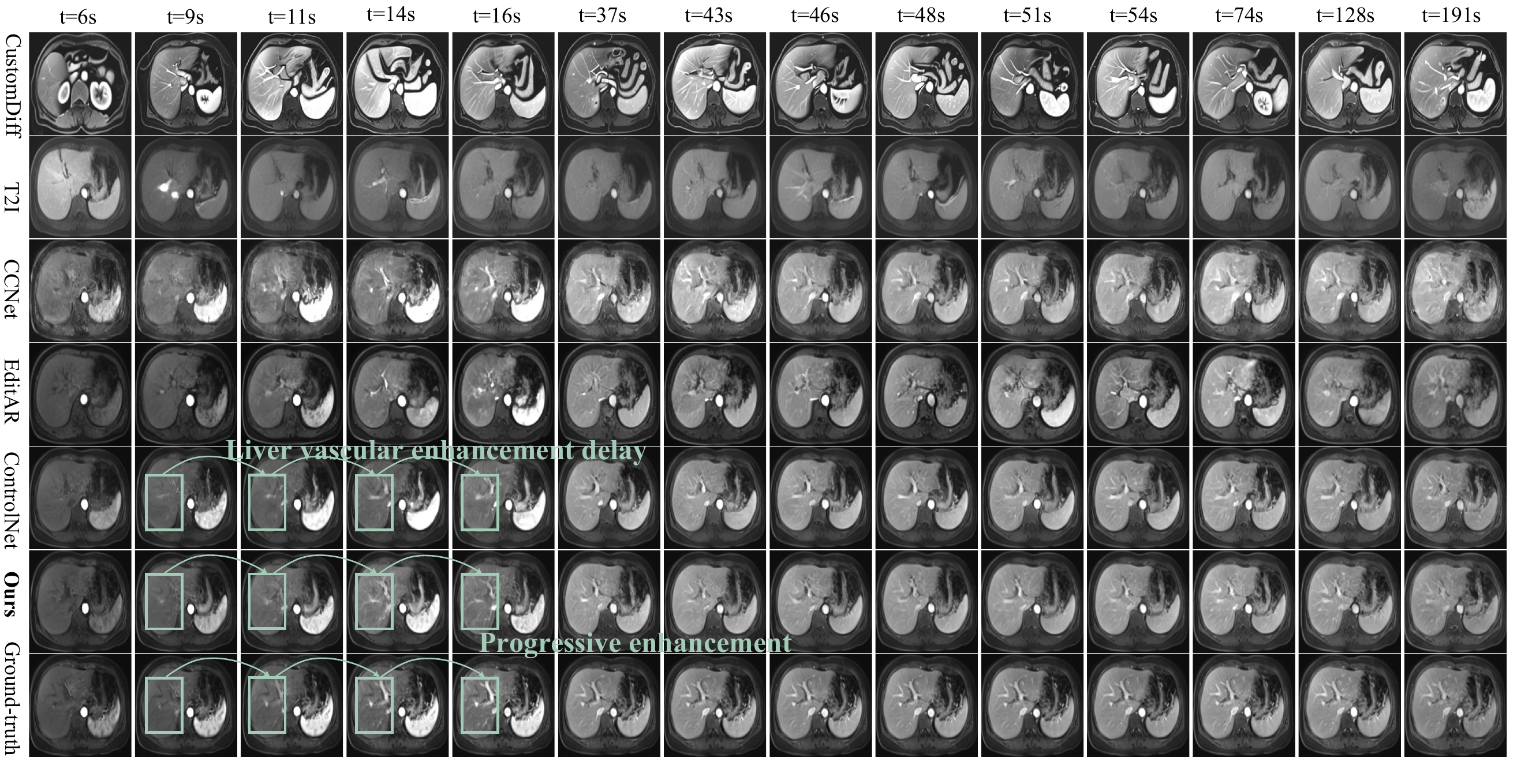}
    \caption{\textbf{Visualization results of liver} in Abdominal DCE-MRI shows that our CEKWorld conforms to the contrast agent kinetic laws of hepatic vessel progressive enhancement and liver parenchyma’s regular enhancement-washout. In contrast, other methods exhibit distorted contents, while ControlNet suffers from non-physiological hepatic vessel enhancement delay, confirming our spatiotemporal consistency learning accurately captures liver contrast agent kinetics.} 
    \label{fig:compare_liver} 
\end{figure*}

\begin{figure*}[h]
    \centering 
    \includegraphics[width=1.0\linewidth]{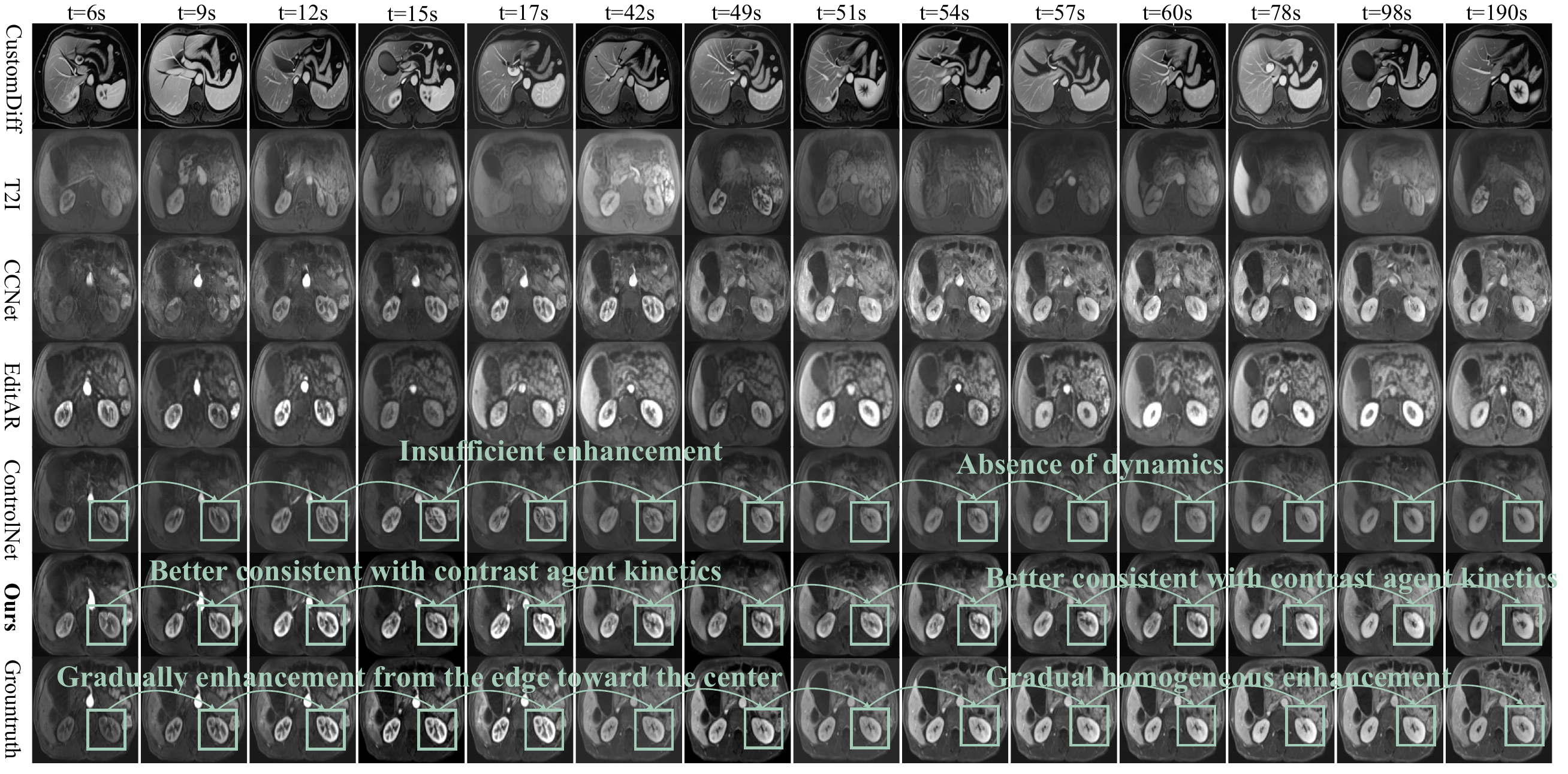}
    \caption{\textbf{Visualization results of kidney} in Abdominal DCE-MRI shows that our CEKWorld reproduces the typical renal perfusion pattern of cortical edge enhancement first, gradual diffusion to the medulla until homogeneity, while methods such as T2I and CCNet have problems of abnormal enhancement or blurred structure, and ControlNet lacks dynamics in the delay phase.} 
    \label{fig:compare_kidney} 
\end{figure*}

\begin{figure*}[h]
    \centering 
    \includegraphics[width=1.0\linewidth]{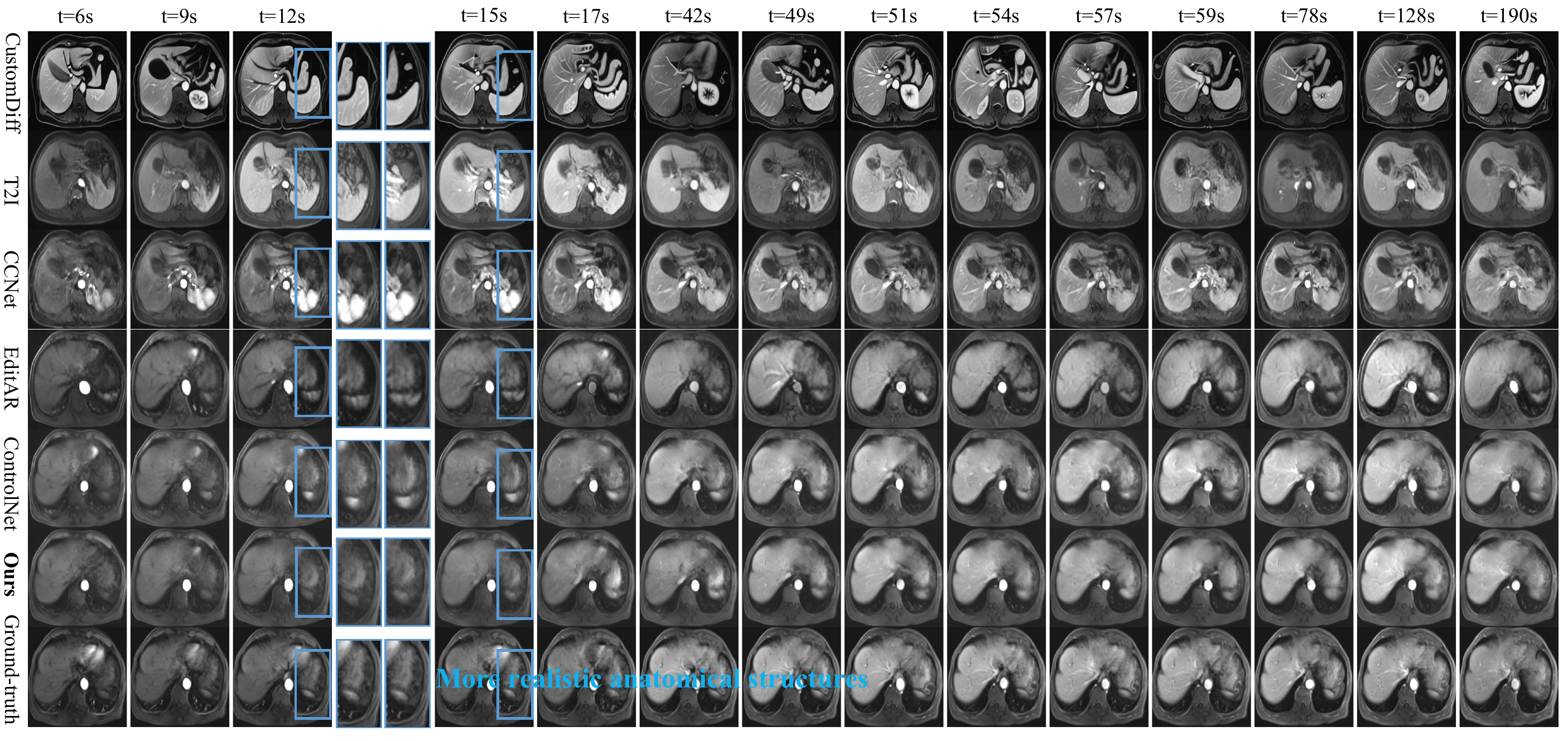}
    \caption{\textbf{Visualization results} in Abdominal DCE-MRI: The visualization results of our methods on different time points exhibit better spatial reality (zoom-in regions in blue boxes).} 
    \label{fig:compare_spatial} 
\end{figure*}

\subsection{Visualization Results}

\subsubsection{Visualization of Different Tissues in Abdominal DCE-MRI}

Different Contrast-enhanced results for the spleen, kidney, and liver demonstrate that CEKWorld is able to simulate different contrast enhancement kinetics laws, which shows the potential for accurate differentiation of normal tissues and lesions across multiple organs, supporting the improvement of diagnostic efficiency and precision in abdominal imaging.

\textit{\textbf{Spleen}} As shown in Fig.\ref{fig:compare_spleen}, CEKWorld conforms to the contrast agent kinetics of the spleen at all time points \cite{mirowitz1991dynamic, elsayes2005mr, vancauwenberghe2015imaging, zhou2013spleen}. In the early arterial phase, when t=6s and 9s, CEKWorld initiates enhancement rapidly and synchronously with the ground-truth, showing typical patchy heterogeneous hyperintensity. In contrast, ControlNet has an obvious delay with slow initiation of enhancement, and methods such as CCNet exhibit abnormal enhancement intensity. In the middle arterial phase, when t=12s and 15s, the enhancement of CEKWorld gradually transitions from patchy to homogeneous, which is completely consistent with the physiological process of contrast agent gradual diffusion in the splenic red pulp sinuses. In contrast, other methods either have distorted enhancement morphology or disordered rhythm. In the late arterial phase and vein phase, when t=17s and 42s, CEKWorld continuously maintains homogeneous hyperintensity without abnormal washout, perfectly matching the imaging feature of the spleen that enhancement lasts long and becomes homogeneous in the later stage. However, other methods have already shown signal distortion or abrupt transition.

\textit{\textbf{Liver}} In Fig.~\ref{fig:compare_liver}, we visualize the predicted dynamic contrast enhancement of the liver across multiple time points. CustomDiff, T2I, CCNet and EditAR all struggle to reproduce realistic liver kinetics \cite{thng2010perfusion, dahlstrom2010quantitative}: vascular structures either fail to enhance at the appropriate phases or exhibit unstable parenchymal signals, with noticeable frame-to-frame fluctuations and over-smoothed textures. ControlNet shows better anatomical fidelity but presents a clear liver vascular enhancement delay—hepatic vessels remain under-enhanced in the early arterial and early portal phases and then suddenly become hyperintense several frames later, resulting in a phase-shifted and non-physiological enhancement pattern. In contrast, our method closely follows the ground truth: the hepatic vessels in the green ROI show progressive enhancement, with smooth, timely transition from arterial to portal venous and delayed phases, while the liver parenchyma brightens and washes out in a gradual, temporally coherent manner. This demonstrates that our spatiotemporal consistency learning not only preserves realistic liver anatomy but also captures the correct contrast-agent kinetics over time.

\textbf{Kidney} In Fig.\ref{fig:compare_kidney}, the contrast agent kinetics in the renal cortex–medulla region highlight that our CEKWorld shows more similarities with the ground-truth sequence. In the ground-truth sequence, enhancement first appears along the cortical rim on the edge, gradually propagates toward the medulla in the center, and finally becomes spatially homogeneous at late phases, exhibiting a typical outside-in renal perfusion pattern \cite{pedersen2021dynamic, eikefjord2015use}. T2I and CCNet either under-enhance or severely corrupt textures, making the cortical–medullary layers indistinguishable; EditAR shows weak and unstable enhancement; and ControlNet suffers from a lack of smooth temporal evolution. In contrast, our CEKWorld reproduces early cortical enhancement, its gradual inward spread, and the final homogeneous enhancement in a structurally faithful and temporally continuous manner, yielding renal contrast dynamics that best match real DCE-MRI.

The virtual angiography visualization results of these tissues conform to the rules of contrast agents, which fully demonstrate that CEKWorld is able to simulate the physiological mechanisms of tissues, the reliability to meet clinical diagnostic needs, as well as the stability and universality across tissue scenarios, and thus has significant clinical application potential.

\subsubsection{Visualization of Spatial Structures in Abdominal DCE-MRI}

In Fig.\ref{fig:compare_spatial}, our methods exhibit better anatomical fidelity. CCNet introduces strong, grainy noise and artifacts in the spleen and liver parenchyma, tearing and blurring organ boundaries, and severely distorting the underlying anatomy. EditAR preserves a roughly correct global outline, but local details fluctuate strongly between neighboring frames: the splenic hilum and adjacent parenchyma alternately swell and collapse, yielding unstable textures. ControlNet largely retains coarse organ shape, yet its internal structures gradually blur, and vessel–parenchyma interfaces drift slightly, indicating mild geometric shift and loss of structural sharpness. In contrast, our MRI CEKWorld consistently preserves spleen contours, liver parenchymal shape, and vascular trajectories that closely match the ground truth. The blue-highlighted regions reveal that our results are the closest to the reference in both organ geometry and texture distribution, demonstrating the effectiveness of our spatial regularization in enforcing patient-specific anatomical consistency.

\begin{figure}[h]
    \centering 
    \includegraphics[width=1.0\linewidth]{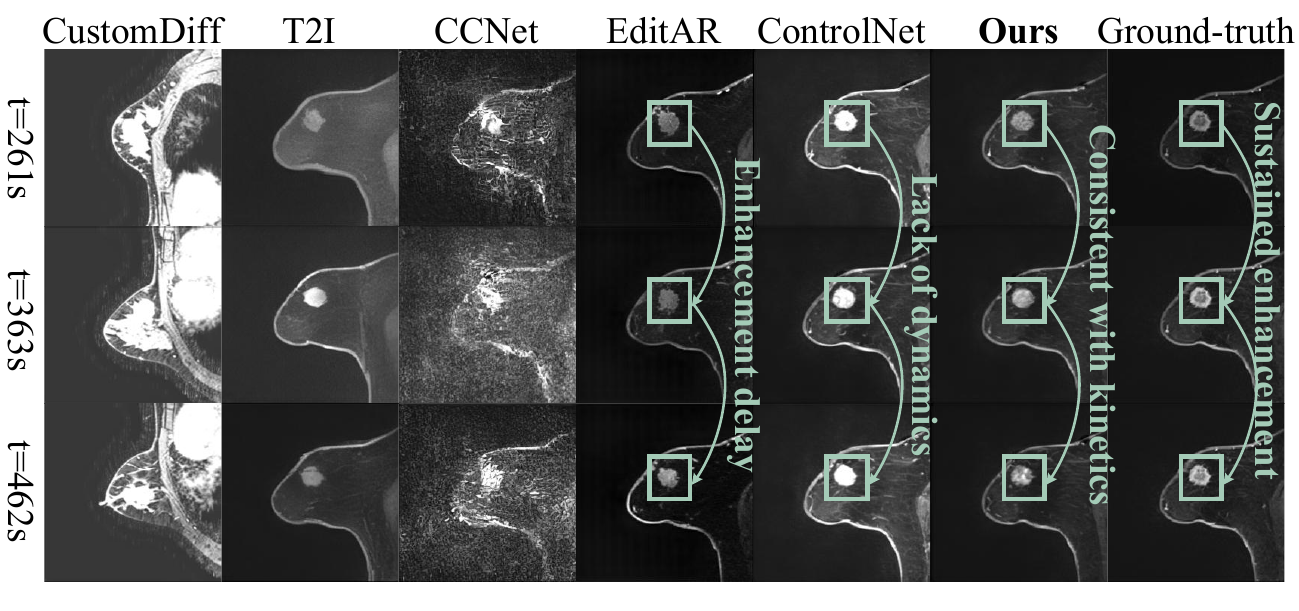}
    \caption{\textbf{Visualization results} in Breast DCE-MRI show that our method exhibits an enhancement pattern similar to the ground truth, characterized by the sustained enhancement, which is often associated with benign or low-risk lesions.} 
    \label{fig:compare_duke1} 
\end{figure}

\begin{figure}[h]
    \centering 
    \includegraphics[width=1.0\linewidth]{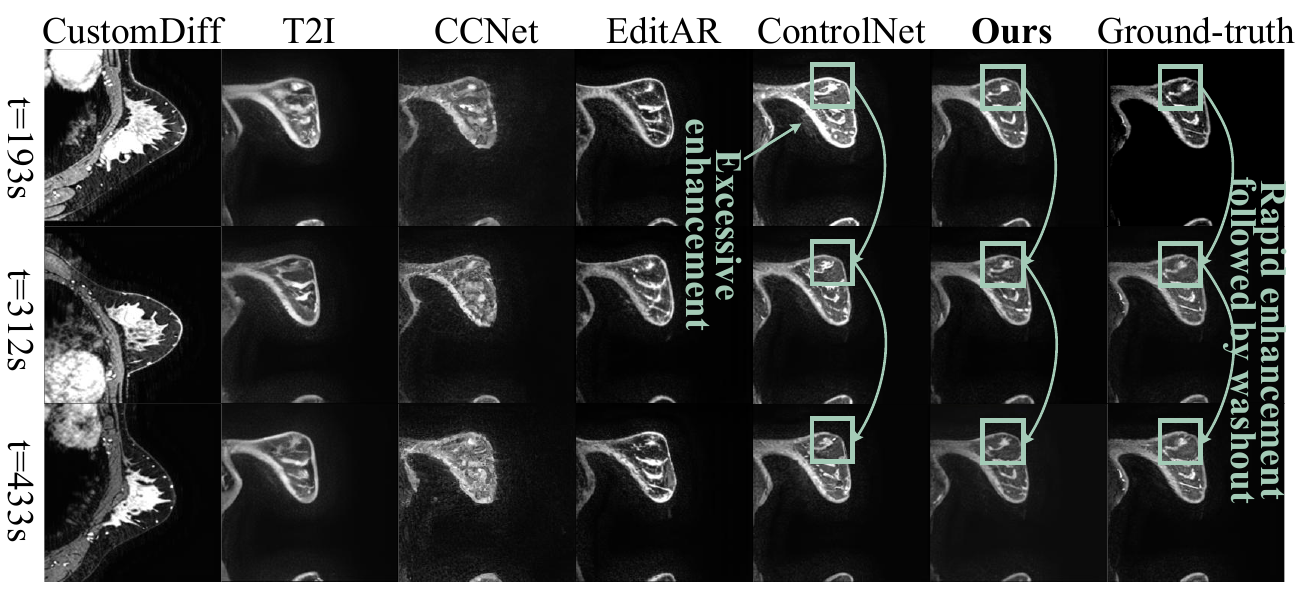}
    \caption{\textbf{Visualization results} in Breast DCE-MRI demonstrate that our method captures the dynamic pattern of rapid initial enhancement followed by clear washout, which is consistent with the ground truth and indicative of malignant or high-risk lesions.} 
    \label{fig:compare_duke2} 
\end{figure}

\subsubsection{Visualization Results for Different Enhancement Patterns in Breast DCE-MRI}
The two breast DCE-MRI visualization cases shown in Fig.\ref{fig:compare_duke1} and Fig.\ref{fig:compare_duke2} demonstrate that our CEKWorld simulates clinically meaningful kinetic patterns, showing strong potential for downstream risk-stratification tasks since lesion malignancy is closely associated with characteristic enhancement patterns \cite{kuhl1999dynamic, macura2006patterns}. 
In both figures, our model accurately reproduces the temporal trajectories observed in ground-truth DCE-MRI sequences, whereas competing methods exhibit severe issues such as incorrect enhancement magnitude, temporal shifts, or lack of temporal evolution.
In Fig.\ref{fig:compare_duke1}, the ground-truth exhibits a gradually rising and persistently elevated curve, typical of a persistent enhancement pattern, often associated with benign or low-risk lesions. Our method produces consistent, sustained enhancement across all time points, faithfully matching the true kinetics. In contrast, EditAR shows enhancement delay, and ControlNet lacks any meaningful temporal progression, failing to reproduce the low-risk kinetic profile. This demonstrates our model’s ability to reconstruct benign-like temporal behavior.
In Fig.\ref{fig:compare_duke2}, the lesion shows rapid initial enhancement followed by clear washout, a hallmark of malignant or high-risk lesions. Our method correctly captures this dynamic pattern: a sharp early rise followed by a decline consistent with the ground truth. EditAR suffers from excessive enhancement and fails to show the washout phase, while ControlNet exhibits collapsed or weakened dynamics, failing to reproduce the malignant-typical kinetic transition.
Together, these two visualization examples show that our method not only preserves spatial fidelity but also faithfully models critical DCE-MRI temporal dynamics. Because low- and high-risk lesions exhibit fundamentally different enhancement trajectories, the ability of our model to reconstruct these dynamics suggests strong potential for downstream tasks such as risk stratification, benign-vs-malignant discrimination, and clinical subtype prediction.